\documentclass[10pt,twocolumn,letterpaper]{article}

\usepackage{cvpr}
\usepackage{times}
\usepackage{epsfig}
\usepackage{graphicx}
\usepackage{amsmath}
\usepackage{amssymb}

\usepackage[numbers,sort]{natbib}
\usepackage{xcolor, colortbl}
\usepackage{tabularx}
\usepackage{multirow}
\usepackage{bbm}
\usepackage{subfiles}

\newcolumntype{C}[1]{>{\centering\let\newline\\\arraybackslash\hspace{0pt}}p{#1}}

\def\ie{\emph{i.e.}}
\def\eg{\emph{e.g.}}

\definecolor{brown}{rgb}{0.65, 0.16, 0.16}
\definecolor{purp}{rgb}{0.65, 0.16, 0.65}
\definecolor{darkgreen}{rgb}{0,0.65,0}
\newcommand{\sh}[1]{{\color{brown}{[#1]}}}

\newcommand{\mcho}[1]{\textcolor{magenta}{#1}}

\usepackage[pagebackref=true,breaklinks=true,letterpaper=true,colorlinks,bookmarks=false]{hyperref}

\cvprfinalcopy 

\def\cvprPaperID{1294} 

\ifcvprfinal\fi
\begin{document}

\title{Deep Metric Learning Beyond Binary Supervision}


\author{
Sungyeon Kim$^\dagger$ \hspace{10mm}
Minkyo Seo$^\dagger$  \hspace{10mm}
Ivan Laptev$^\ddagger$  \hspace{10mm}
Minsu Cho$^\dagger$    \hspace{10mm}
Suha Kwak$^\dagger$\\
POSTECH, Pohang, Korea$^\dagger$ \hspace{10mm} 
Inria / \'Ecole Normale Sup\'erieure, Paris, France$^\ddagger$ \\
{\tt\small \{tjddus9597, mkseo, mscho, suha.kwak\}@postech.ac.kr, ivan.laptev@inria.fr}
}

\maketitle


\begin{abstract}
Metric Learning for visual similarity has mostly adopted binary supervision indicating whether a pair of images are of the same class or not.
Such a binary indicator covers only a limited subset of image relations, and is not sufficient to represent semantic similarity between images described by continuous and/or structured labels such as object poses, image captions, and scene graphs.
Motivated by this, we present a novel method for deep metric learning using continuous labels.
First, we propose a new triplet loss that allows distance ratios in the label space to be preserved in the learned metric space.
The proposed loss thus enables our model to learn the degree of similarity rather than just the order.
Furthermore, we design a triplet mining strategy adapted to metric learning with continuous labels.
We address three different image retrieval tasks with continuous labels in terms of human poses, room layouts and image captions, and demonstrate the superior performance of our approach compared to previous methods.

\end{abstract}


\section{Introduction}
\label{sec:intro}

%
The sense of similarity has been known as the most basic component of human reasoning~\cite{Quine1969}.
Likewise, understanding similarity between images has played essential roles in many areas of computer vision including 
image retrieval~\cite{Wang2014,songCVPR17,songCVPR16,PDDM}, 
face identification~\cite{Schroff2015,Chopra2005,multibatch_embedding}, 
place recognition~\cite{NetVLAD}, 
pose estimation~\cite{Sumer_2017_ICCV}, 
person re-identification~\cite{Shi_ECCV_2016,Chen_2017_CVPR}, 
video object tracking~\cite{siamese_tracker,Son_2017_CVPR}, 
local feature descriptor learning~\cite{Kumar_CVPR_2016,Zagoruyko_CVPR_2015}, 
zero-shot learning~\cite{Bucher_ECCV_2016,Yuan_2017_ICCV}, 
and self-supervised representation learning~\cite{Wang2015}.
Also, the perception of similarity has been achieved by learning similarity metrics from labeled images, which is called \emph{metric learning}. 

\begin{figure} [!t]
\centering
\includegraphics[width = 0.97 \columnwidth]{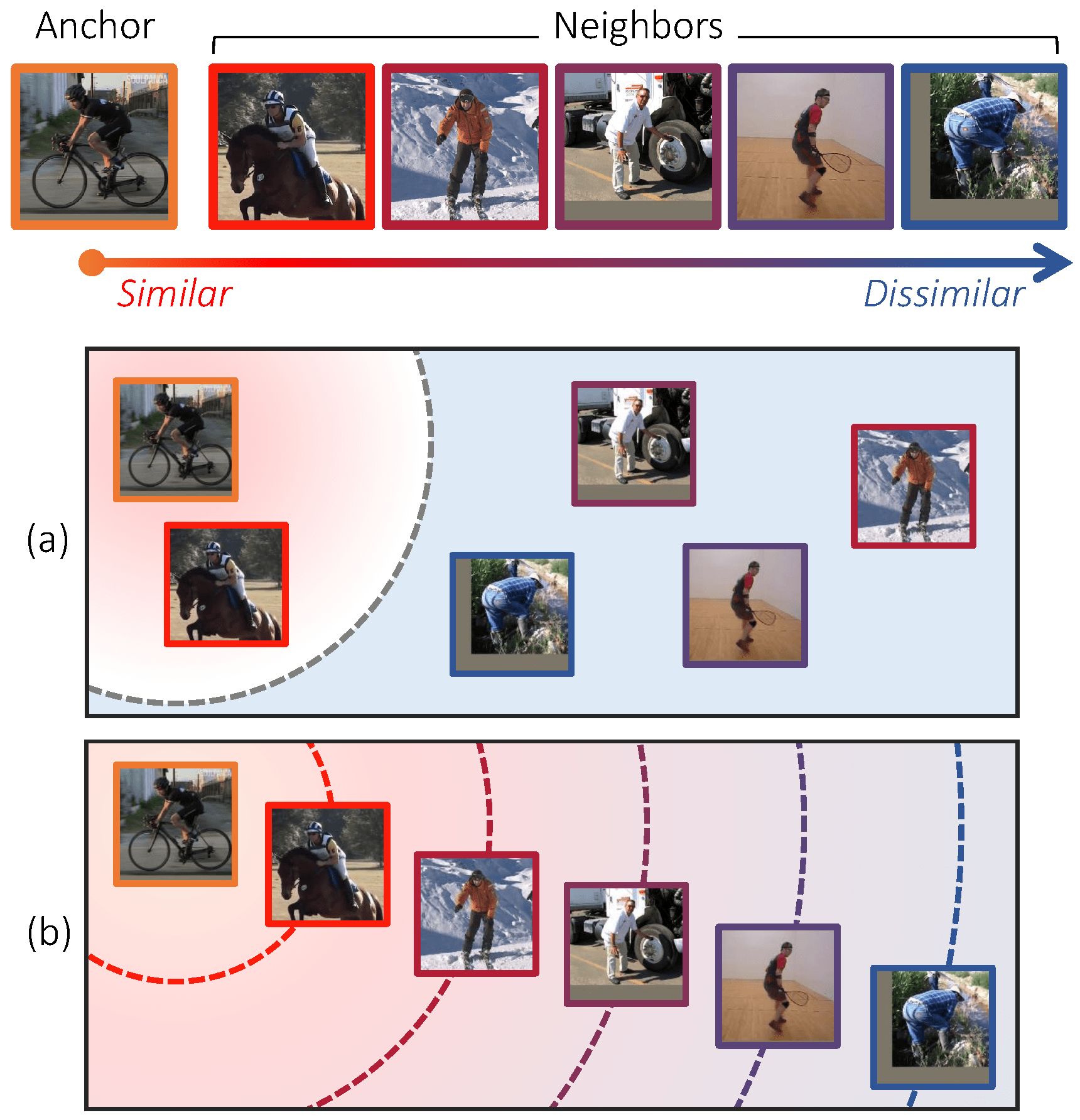}
\caption{
A conceptual illustration for comparing existing methods~\cite{NetVLAD,Mori2015,Sumer_2017_ICCV,thin_slicing,Gordo_cvpr2017} and ours.
Each image is labeled by human pose, and colored in red if its pose similarity to the anchor is high. 
(a) Existing methods categorize neighbors into positive and negative classes, and learn a metric space where positive images are close to the anchor and negative ones far apart.
In such a space, the distance between a pair of images is not necessarily related to their semantic similarity since the order and degrees of similarities between them are disregarded.
(b) Our approach allows distance ratios in the label space to be preserved in the learned metric space so as to overcome the aforementioned limitation.
} 
\label{fig:motivation}
\end{figure}

Recent approaches in metric learning have improved performance dramatically by adopting deep Convolutional Neural Networks (CNNs) as their embedding functions.
Specifically, such methods train CNNs to project images onto a manifold where two examples are close to each other if they are semantically similar and far apart otherwise. 
While in principle such a metric can be learned using any type of semantic similarity labels, previous approaches typically rely on binary labels over image pairs indicating whether the image pairs are similar or not.
In this aspect, only a small subset of real-world image relations has been addressed by previous approaches.
Indeed, binary similarity labels are not sufficient to represent sophisticated
relations between images with structured and continuous labels, such as
image captions~\cite{Mscoco,Flickr30k_a,Flickr30k_b}, human poses~\cite{MPII_pose_dataset,Johnson2011}, camera poses~\cite{relocnet,ScanNet}, and scene graphs~\cite{visual_genome,visual_relation_detection}. 
Metric learning with continuous labels has been addressed in~\cite{thin_slicing,Mori2015,Sumer_2017_ICCV,Gordo_cvpr2017,NetVLAD}.
Such methods, however, reduce the problem by quantizing continuous similarity into binary labels (\ie, \emph{similar} or \emph{dissimilar}) and applying the existing metric learning techniques.
Therefore, they do not fully exploit rich similarity information in images with continuous labels as illustrated in Figure~\ref{fig:motivation}(a) and require a careful tuning of parameters for the quantization.

In this paper, we propose a novel method for deep metric learning to overcome the aforementioned limitations.
We first design a new triplet loss function that takes full advantage of continuous labels in metric learning.
Unlike existing triplet losses~\cite{LMNN,Schroff2015,Wohlhart2015} that are interested only in the equality of class labels or the order of label distances, our loss aims to preserve ratios of label distances in the learned embedding space. 
This allows our model to consider degrees of similarities as well as their order and to capture richer similarity information between images as illustrated in Figure~\ref{fig:motivation}(b).

Current methods construct triplets by sampling a positive (\emph{similar}) and a negative (\emph{dissimilar}) examples to obtain the binary supervision. 
Here we propose a new strategy for triplet sampling. Given a minibatch composed of an anchor and its neighbors, our method samples every triplet including the anchor by choosing every pair of neighbors in the minibatch. 
Unlike the conventional approaches, our method does not need to introduce quantization parameters to categorize neighbors into the two classes and can utilize more triplets given the same minibatch.

Our approach can be applied to various problems with continuous and structured labels.
We demonstrate the efficacy of the proposed method on three different image retrieval tasks using human poses, room layouts, and image captions, respectively, as continuous and structured labels.
In all the tasks, our method outperforms the state of the art, and our new loss and the triplet mining strategy both contribute to the performance boost.
Moreover, we find that our approach learns a better metric space even with a significantly lower embedding dimensionality compared to previous ones.
Finally, we show that a CNN trained by our method with caption similarity can serve as an effective visual feature for image captioning, and it outperforms an ImageNet pre-trained counterpart in the task.

\begin{figure*} [!t]
\centering
\includegraphics[width = 0.99 \textwidth]{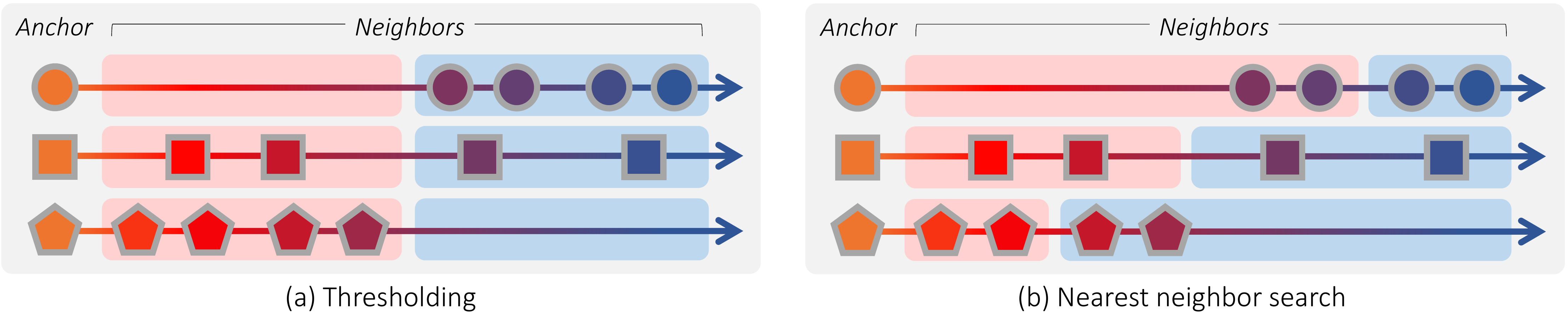}
\caption{
The binary quantization strategies and their limitations. 
The orange circle indicates a rare example dissimilar to most of the others, and the orange pentagon is a common example similar with a large number of samples.
(a) If the quantization is done by a single distance threshold, 
populations of positive and negative examples would be significantly imbalanced.
(b) In the case of nearest neighbor search, 
positive neighbors of a rare example would be dissimilar and negative neighbors of a common example would be too similar.
} 
\label{fig:quantization}
\end{figure*}

\section{Related Work}
\label{sec:relatedwork}

In this section, we first review loss functions and tuple mining techniques for deep metric learning, then discuss previous work on metric learning with continuous labels.

\subsection{Loss Functions for Deep Metric Learning}
\label{sec:prev_loss}

Contrastive loss~\cite{Chopra2005,Hadsell2006,Bromley1994} and triplet loss~\cite{Schroff2015,Wohlhart2015,Wang2014} are standard loss functions for deep metric learning.
Given an image pair, the contrastive loss minimizes their distance in the embedding space if their classes are the same, and separates them a fixed margin away otherwise. 
The triplet loss takes triplets of anchor, positive, and negative images, and enforces the distance between the anchor and the positive to be smaller than that between the anchor and the negative. 
One of their extensions is quadruple loss~\cite{Son_2017_CVPR,Chen_2017_CVPR}, which considers relations between a quadruple of images and is formulated as a combination of two triplet losses.
A natural way to generalize the above losses is to use a higher order relations.
For example, $n$-tuplet loss~\cite{Sohn_nips2016} takes as its input an anchor, a positive, and $n-2$ negative images, and jointly optimizes their embedding vectors.
Similarly, lifted structured loss~\cite{songCVPR16} considers all positive and negative pairs in a minibatch at once by incorporating hard-negative mining functionality within itself.
For the same purpose, in~\cite{Ustinova_NIPS_2016} the area of intersection between similarity distributions of positive and negative pairs are minimized, and in~\cite{songCVPR17,Law_ICML17} clustering objectives are adopted for metric learning.

All the aforementioned losses utilize image-level class labels or their equivalent as supervision. 
Thus, unlike ours, it is not straightforward for them to take relations between continuous and/or structured labels of images into account.

\subsection{Techniques for Mining Training Tuples}
\label{sec:tuple_mining}
Since tuples of $k$ images are used in training, the number of possible tuples increases exponentially with $k$.
The motivation of mining techniques is that some of such a large number of tuples do not contribute to training or can even result in decreased performance.
A representative example is semi-hard triplet mining~\cite{Schroff2015}, which utilizes only semi-hard triplets for training since easy triplets do not update the network and hardest ones may have been corrupted due to labeling errors.
It also matters how to measure the hardness. 
A common strategy~\cite{Schroff2015,songCVPR16} is to utilize pairwise Euclidean distances in embedding space, \eg, negative pairs with small Euclidean distances are considered hard.
In~\cite{sampling_matters,PDDM,Iscen_2018_CVPR}, an underlying manifold of embedding vectors, which is ignored in Euclidean distances, is taken into account to improve the effectiveness of mining techniques.
Also, in~\cite{Yuan_2017_ICCV} multiple levels of hardness are captured by a set of embedding models with different complexities.


Although the above techniques substantially improve the quality of learned embedding space, they are commonly based on binary relations between image pairs, thus they are not directly applicable for metric learning with continuous labels.

\subsection{Metric Learning Using Continuous Labels}
\label{sec:prev_cont_ml}

There have been several metric learning methods using data with continuous labels. 
For example, similarities between human pose annotations have been used to learn an image embedding CNN~\cite{thin_slicing,Mori2015,Sumer_2017_ICCV}.
This pose-aware CNN then extracts pose information of given image efficiently without explicit pose estimation, which can be transferred to other tasks relying on pose understanding like action recognition.
Also, in~\cite{Gordo_cvpr2017} caption similarities between image pairs are used as labels for metric learning, and the learned embedding space enables image retrieval based on more comprehensive understanding of image content.
Other examples of continuous labels that have been utilized for metric learning include GPS data for place recognition~\cite{NetVLAD} and camera frusta for camera relocalization~\cite{relocnet}.

However, it is hard for the above methods to take full advantage of continuous labels because they all use conventional metric learning losses based on binary relations.
Due to their loss functions, they quantize continuous similarities into binary levels through distance thresholding~\cite{NetVLAD,Mori2015,Sumer_2017_ICCV} or nearest neighbor search~\cite{thin_slicing,Gordo_cvpr2017}.
Unfortunately, both strategies are unnatural for continuous metric learning and have clear limitations as illustrated in Figure~\ref{fig:quantization}.
Furthermore, it is not straightforward to find a proper value for their quantization parameters since there is no clear boundary between positive and negative examples whose distances to the anchors are continuous.
To the best of our knowledge, our work is the first attempt to {\em directly} use continuous labels for metric learning.

\section{Our Framework}
\label{sec:method}

To address limitations of existing methods described above, we propose a new triplet loss called \emph{log-ratio loss}.
Our loss directly utilizes continuous similarities without quantization.
Moreover, it considers degrees of similarities as well as their rank so that the resulting model can infer sophisticated similarity relations between continuous labels. 
In addition, we present a new, simple yet effective triplet mining strategy supporting our log-ratio loss since the existing mining techniques in Section~\ref{sec:tuple_mining} cannot be used together with our loss.

In the following sections, we briefly review the conventional triplet loss~\cite{Schroff2015} for a clear comparison, then present details of our log-ratio loss and the new triplet mining technique.

\subsection{Review of Conventional Triplet Loss}
The triplet loss takes a triplet of an anchor, a positive, and a negative image as input.
It is designed to penalize triplets violating the rank constraint, namely, that the distance between the anchor and the positive must be smaller than that between the anchor and the negative in the embedding space.
The loss is formulated as
\begin{equation}
\ell_{\textrm{tri}}(a,p,n) = \Big[ D(f_a, f_p) - D(f_a, f_n) + \delta \Big]_+, \label{eq:triplet}
\end{equation}
where $f$ indicates an embedding vector, $D(\cdot)$ means the squared Euclidean distance, $\delta$ is a margin, and $[\cdot]_+$ denotes the hinge function.
Note that the embedding vectors should be $L_2$ normalized since, without such a normalization, their magnitudes tend to diverge and the margin becomes trivial.
For training, gradients of $\ell_{\textrm{tri}}$ with respect to the embedding vectors are computed by
\begin{align}
\frac{\partial \ell_{\textrm{tri}}(a,p,n)}{\partial f_p} & = 2 (f_p - f_a) \cdot \mathbbm{1}\big( \ell_{\textrm{tri}}(a,p,n) > 0 \big ), \label{eq:triplet_grad_p} \\
\frac{\partial \ell_{\textrm{tri}}(a,p,n)}{\partial f_n} & = 2 (f_a - f_n) \cdot \mathbbm{1}\big( \ell_{\textrm{tri}}(a,p,n) > 0 \big ), \label{eq:triplet_grad_n} \\
\frac{\partial \ell_{\textrm{tri}}(a,p,n)}{\partial f_a} & = - \frac{\partial \ell_{\textrm{tri}}(a,p,n)}{\partial f_p} - \frac{\partial \ell_{\textrm{tri}}(a,p,n)}{\partial f_n}, \label{eq:triplet_grad_a}
\end{align}
where $\mathbbm{1}$ is the indicator function.
One may notice that the gradients only consider the directions between the embedding vectors and the rank constraint violation indicator. 
If the rank constraint is satisfied, all the gradients are zero.

\subsection{Log-ratio Loss}
Given a triplet with samples, we propose a log-ratio loss that aims to approximate the ratio of label distances by the ratio of distances in the learned embedding space.
Specifically, we define the loss function as
\begin{equation}
\begin{split}
    \ell_{\textrm{lr}}(a,i,j) =
    \bigg\{
        \log{\frac{D(f_a,f_i)}{D(f_a,f_j)}}
        -\log{\frac{D(y_a,y_i)}{D(y_a,y_j)}}
    \bigg\}^2, \label{eq:log_ratio}
\end{split}
\end{equation}
where $f$ indicates an embedding vector, $y$ is a continuous label, and $D(\cdot)$ denotes the squared Euclidean distance.
Also, $(a,i,j)$ is a triplet of an anchor $a$ and its two neighbors $i$ and $j$ without positive-negative separation, unlike $p$ and $n$ in Eq.~\eqref{eq:triplet}.
By approximating ratios between label distances instead of the distances themselves, the proposed loss enables to learn a metric space more flexibly regardless of the scale of the labels.


The main advantage of the log-ratio loss is that it allows a learned metric space to reflect degrees of label similarities as well as the rank of them.
Ideally, the distance between two images in the learned metric space will be proportional to their distance in the label space.
Hence, an embedding network trained with our loss can represent continuous similarities between images more thoroughly than those focusing only on the rank of similarities like the triplet loss.
This property of the log-ratio loss can be also explained through its gradients, which are given by 
\begin{align}
\frac{\partial \ell_{\textrm{lr}}(a,i,j)}{\partial f_i} & = \frac{(f_i - f_a)}{D(f_a,f_i)} \cdot \ell'_{\textrm{lr}}(a,i,j), \label{eq:logratio_grad_i} \\
\frac{\partial \ell_{\textrm{lr}}(a,i,j)}{\partial f_j} & = \frac{(f_a - f_j)}{D(f_a,f_j)} \cdot \ell'_{\textrm{lr}}(a,i,j), \label{eq:logratio_grad_j} \\
\frac{\partial \ell_{\textrm{lr}}(a,i,j)}{\partial f_a} & = - \frac{\partial \ell_{\textrm{lr}}(a,i,j)}{\partial f_i} - \frac{\partial \ell_{\textrm{lr}}(a,i,j)}{\partial f_j}, \label{eq:logratio_grad_a}
\end{align}
where $\ell'_{\textrm{lr}}(a,i,j)$ is a scalar value computed by
\begin{equation}
\ell'_{\textrm{lr}}(a,i,j) = 4 \left\{ \log{\frac{D(f_a,f_i)}{D(f_a,f_j)}} - \log{\frac{D(y_a,y_i)}{D(y_a,y_j)}} \right\}. \label{eq:log-ratio-discrepancy}
\end{equation}
As shown in Eq.~\eqref{eq:logratio_grad_i} and~\eqref{eq:logratio_grad_j}, the gradients of the log-ratio loss are determined not only by the directions between the embedding vectors but also by $\ell'_{\textrm{lr}}(a,i,j)$ that quantifies the discrepancy between the distance ratio in the label space and that in the embedding space.
Thus, even when the rank constraint is satisfied, the magnitudes of the gradients could be significant if $\ell'_{\textrm{lr}}(a,i,j)$ is large.
In contrast, the gradients of the triplet loss in Eq.~\eqref{eq:triplet_grad_p} and~\eqref{eq:triplet_grad_n} become zero under the same condition. 


Another advantage of the log-ratio loss is that it is parameter-free. 
Unlike ours, the triplet loss requires the margin, which is a hyper-parameter tuned manually and forces embedding vectors to be $L_2$ normalized.
Last but not least, we empirically find that the log-ratio loss can outperform the triplet loss even with embeddings of a significantly lower dimensionality, which enables a more efficient and effective image retrieval.

\subsection{Dense Triplet Mining}
The existing triplet mining methods in Section~\ref{sec:tuple_mining} cannot be used in our framework since they are specialized to handle images annotated by discrete and categorical labels.
Hence, we design our own triplet mining method that is well matched with the log-ratio loss.


First of all, we construct a minibatch $B$ of training samples with an anchor, $k$ nearest neighbors of the anchor in terms of label distance, and other neighbors randomly sampled from the remaining ones. 
Note that including nearest neighbors helps speed up training.
Since the label distance between an anchor and its nearest neighbor is relatively small, triplets with a nearest neighbor sample in general induce large log-ratios of label distances in Eq.~\eqref{eq:log-ratio-discrepancy}, which may increase the magnitudes of the associated gradients consequently.

Given a minibatch, we aim to exploit all triplets sharing the anchor so that our embedding network can observe the greatest variety of triplets during training.
To this end, we sample triplets by choosing every pair of neighbors $(i,j)$ in the minibatch and combining them with the anchor $a$.
Furthermore, since $(a,i,j)$ and $(a,j,i)$ have no difference in our loss, we choose only $(a,i,j)$ and disregard $(a,j,i)$ when $D(y_a, y_i) < D(y_a, y_j)$ to avoid duplication.
We call the above procedure \emph{dense triplet mining}.
The set of triplets densely sampled from the minibatch $B$ is then given by
\begin{align}
\mathcal{T}(B) = \big\{(a, i, j) \ | & \ D(y_a, y_i) < D(y_a, y_j), \\
                                     & \ i \in B \setminus \{a\}, \ j \in B \setminus \{a\} \big\}. \nonumber
\end{align}


Note that our dense triplet mining strategy can be combined also with the triplet loss, which is re-formulated as
\begin{eqnarray}
&\ell_{\textrm{tri}}^{\textrm{dense}}(a,i,j) = \Big[ D(f_a, f_i) - D(f_a, f_j) + \delta \Big]_+ & \label{eq:triplet_dense} \\
&\textrm{subject to } \ (a,i,j)\in \mathcal{T}(B).& \nonumber \label{eq:dense_tri}
\end{eqnarray}
where the margin $\delta$ is set small compared to that of $\ell_{\textrm{tri}}$ in Eq.~\eqref{eq:triplet} since the label distance between $i$ and $j$ could be quite small when they are densely sampled.
This dense triplet loss is a strong baseline of our log-ratio loss.
However, it still requires $L_2$ normalization of embedding vectors and ignores degrees of similarities as the conventional triplet loss does.
Hence, it can be regarded as an intermediary between the existing approaches in Section~\ref{sec:prev_cont_ml} and our whole framework, and will be empirically analyzed for ablation study in the next section.
\section{Experiments}
\label{sec:experiments}

The effectiveness of the proposed framework is validated on three different image retrieval tasks based on continuous similarities: human pose retrieval on the MPII human pose dataset~\cite{MPII_pose_dataset}, room layout retrieval on the LSUN dataset~\cite{zhang2015large}, and caption-aware image retrieval on the MS-COCO dataset~\cite{Mscoco}.
We also demonstrate that an image embedding CNN trained with caption similarities through our framework can be transferred to image captioning 
as an effective visual representation.

In the rest of this section, we first define evaluation metric and describe implementation details, then present qualitative and quantitative analysis of our approach on the retrieval and representation learning tasks.



\subsection{Evaluation: Measures and Baselines}
\noindent \textbf{Evaluation metrics.}
Since image labels are continuous and/or structured in our retrieval tasks, 
it is not appropriate to evaluate performance based on standard metrics like Recall@$k$.
Instead, following the protocol in~\cite{thin_slicing}, we adopt two evaluation metrics, mean label distance and a modified version of nDCG~\cite{Burges2005,thin_slicing}.
The mean label distance is the average of distances between queries and retrieved images in the label space, and a smaller means a better retrieval quality.
The modified nDCG considers the rank of retrieved images as well as their relevance scores, and is defined as
\begin{equation}
\begin{split}
    \textrm{nDCG}_K(q) = \frac{1}{Z_K}\sum^K_{i=1}{\frac{2^{r_i}}{\log_2{(i+1)}}},
\end{split}
\end{equation}
where $K$ is the number of top retrievals of our interest and $Z_K$ is a normalization factor to guarantee that the maximum value of $\textrm{nDCG}_K$ is 1.
Also, $r_i = -\log_2{(\lVert{y_q - y_i}\lVert_2+1)} $ denotes the relevance between query $q$ and the $i^{\textrm{th}}$ retrieval, which is discounted by $\log_2{(i+1)}$ to place a greater emphasis on one returned at a higher rank.
A higher nDCG means a better retrieval quality.

\noindent \textbf{Common baselines.}
In the three retrieval tasks, our method is compared with its variants for ablation study.
These approaches are denoted by combinations of loss function $L$ and triplet mining strategy $M$, where \textsf{\small Log-ratio} is our log-ratio loss, \textsf{\small Triplet} means the triplet loss, \textsf{\small Dense} denotes the dense triplet mining, and \textsf{\small Binary} indicates the triplet mining based on binary quantization.
Specifically, $M$(\textsf{\small Binary}) is implemented by nearest neighbor search, where 30 neighbors closest to anchor are regarded as positive.
Our model is then represented as $L$(\textsf{\small Log-ratio})+$M$(\textsf{\small Dense}).
We also compare our model with the same network trained with the margin based loss and distance weighted sampling~\cite{sampling_matters}, a state-of-the-art approach in conventional metric learning.
Finally, we present scores of \textsf{\small Oracle} and ImageNet pretrained ResNet-34 as upper and lower performance bounds.
Note that nDCG of \textsf{\small Oracle} is always 1.


\subsection{Implementation Details}
\noindent \textbf{Datasets.}
For the human pose retrieval,
we directly adopt the dataset and setting of~\cite{thin_slicing}.
Among in total 22,285 full-body pose images, 12,366 images are used for training and 9,919 for testing, while 1,919 images among the test set are used as queries for retrieval.
For the room layout retrieval, we adopt the LSUN room layout dataset~\cite{zhang2015large} that contains 4,000 training images and 394 validation images of 11 layout classes. 
Since we are interested in continuous and fine-grained labels only, 
we use only 1,996 images of the 5$^\textrm{th}$ layout class, which is the class with the largest number of images.
Among them 1,808 images are used for training and 188 for testing, in which 30 images are employed as queries.
Finally, for the caption-aware image retrieval, the MS-COCO 2014 caption dataset~\cite{Mscoco} is used. 
We follow the Karpathy split~\cite{Karpathy_mscoco_split}, where 113,287 images are prepared for training and 5,000 images for validation and testing, respectively.
The retrieval test is conducted only on the testing set, where 500 images are used as queries.

\noindent \textbf{Preprocessing and data augmentation.}
For the human pose retrieval, we directly adopt the data augmentation techniques used in~\cite{thin_slicing}.
For the room layout retrieval, the images are resized to $224 \times 224$ for both training and testing, and flipped horizontally at random during training. 
For the caption-aware retrieval, images are jittered in both scale and location, cropped to $224 \times 224$, and flipped horizontally at random during training.
Meanwhile, test images are simply resized to $256 \times 256$ and cropped at center to $224 \times 224$. 

\noindent \textbf{Embedding networks and their training.}
For the human pose and room layout retrieval, we choose ResNet-34~\cite{resnet} as our backbone network and append a 128-D FC layer on top for embedding.
They are optimized by the SGD with learning rate $10^{-2}$ and exponential decay for 15 epochs.
For the caption-aware image retrieval, ResNet-101~\cite{resnet} with a 1,024 dimensional embedding layer is adopted since captions usually contain more comprehensive information than human poses and room layouts.
This network is optimized by the ADAM~\cite{Adamsolver} with learning rate $5\cdot 10^{-6}$ for 5 epochs.
All the networks are implemented in PyTorch~\cite{pytorch} and pretrained on ImageNet~\cite{Russakovsky2015} before being finetuned. 


\noindent \textbf{Hyper-parameters.}
The size of minibatch is set to 150 for the human pose, 100 for the room layout, and 50 for the caption-aware image retrieval, respectively.
On the other hand, $k$, the number of nearest neighbors in the minibatch for the dense triplet mining, is set to 5 for all experiments.
For the common baselines, 
the margin $\delta$ of the conventional triplet loss is set to 0.2 and
that of the dense triplet loss 0.03.

\begin{figure*} [!t]
\centering
\includegraphics[width = 1 \textwidth]{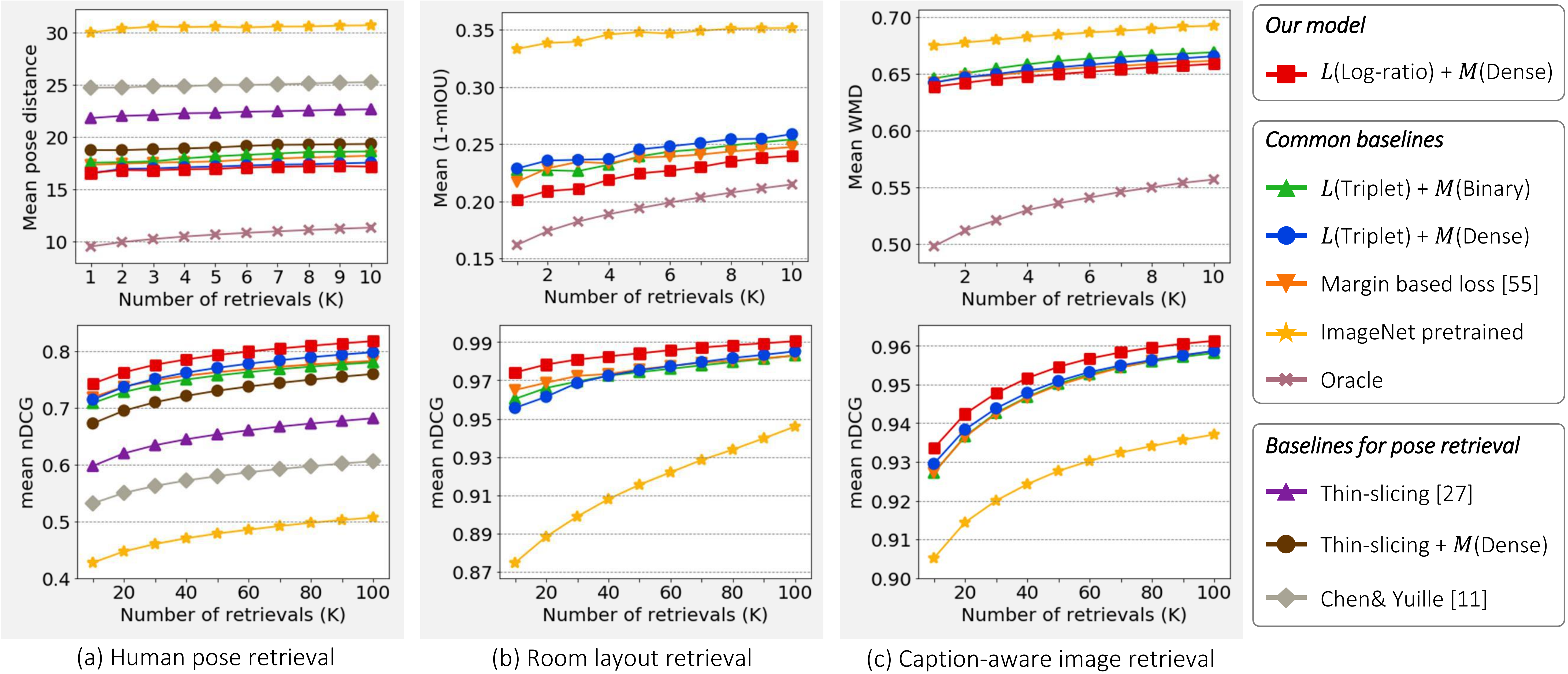}
\caption{
Quantitative evaluation of the three retrieval tasks in terms of mean label distance (\emph{top}) and mean nDCG (\emph{bottom}). 
} 
\vspace{-0.2cm}
\label{fig:quantitative_eval}
\end{figure*}

\begin{figure} [t]
\centering
\includegraphics[width = 1 \columnwidth]{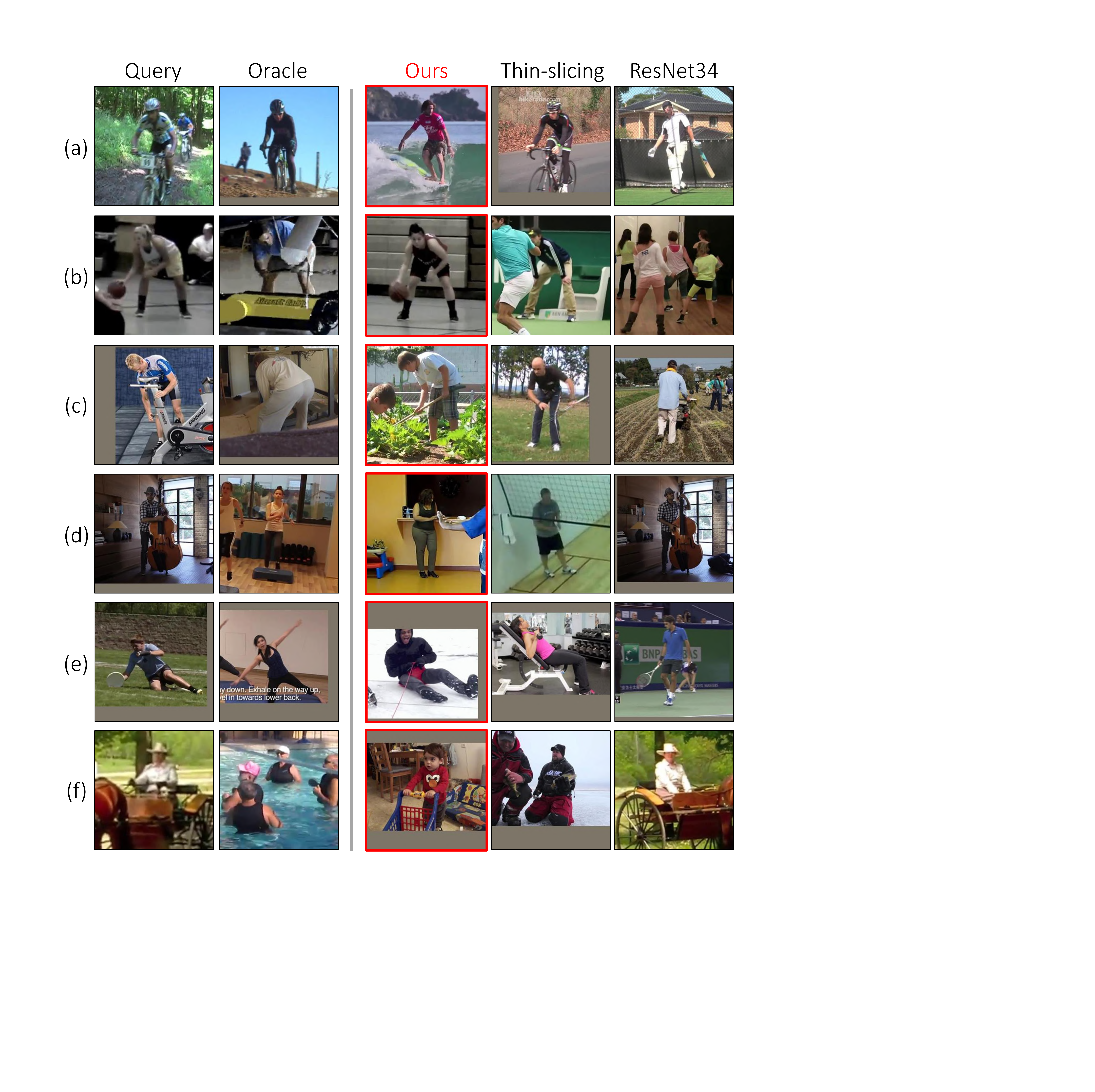}
\caption{
Qualitative results of human pose retrieval.
} 
\vspace{-0.3cm}
\label{fig:pose_qualitative}
\end{figure}

\subsection{Human Pose Retrieval}
The goal of human pose retrieval is to search for images similar with query in terms of human poses they exhibit.
Following~\cite{thin_slicing}, the distance between two poses is defined as the sum of Euclidean distances between body-joint locations. 
Our model is compared with the previous pose retrieval model called thin-slicing~\cite{thin_slicing} and a CNN for explicit pose estimation~\cite{Chen2014} as well as the common baselines.

Quantitative evaluation results of these approaches are summarized in Figure~\ref{fig:quantitative_eval}(a), where our model clearly outperforms all the others.
In addition, through comparisons between ours and its two variants 
$L$(\textsf{\small Triplet})+$M$(\textsf{\small Dense}) and 
$L$(\textsf{\small Triplet})+$M$(\textsf{\small Binary}),
it is demonstrated that both of our log-ratio loss and the dense triplet mining contribute to the improvement.
Qualitative examples of human pose retrieval are presented in Figure~\ref{fig:pose_qualitative}.
Our model and thin-slicing overall successfully retrieve images exhibiting similar human poses with queries, while ResNet-34 focuses mostly on object classes and background components. 
Moreover, ours tends to capture subtle characteristics of human poses (\eg, bending left-arms in Figure~\ref{fig:pose_qualitative}(b))
and handle rare queries (\eg, Figure~\ref{fig:pose_qualitative}(e)) 
better than thin-slicing.

\begin{figure} [!t]
\centering
\includegraphics[width = 1 \columnwidth]{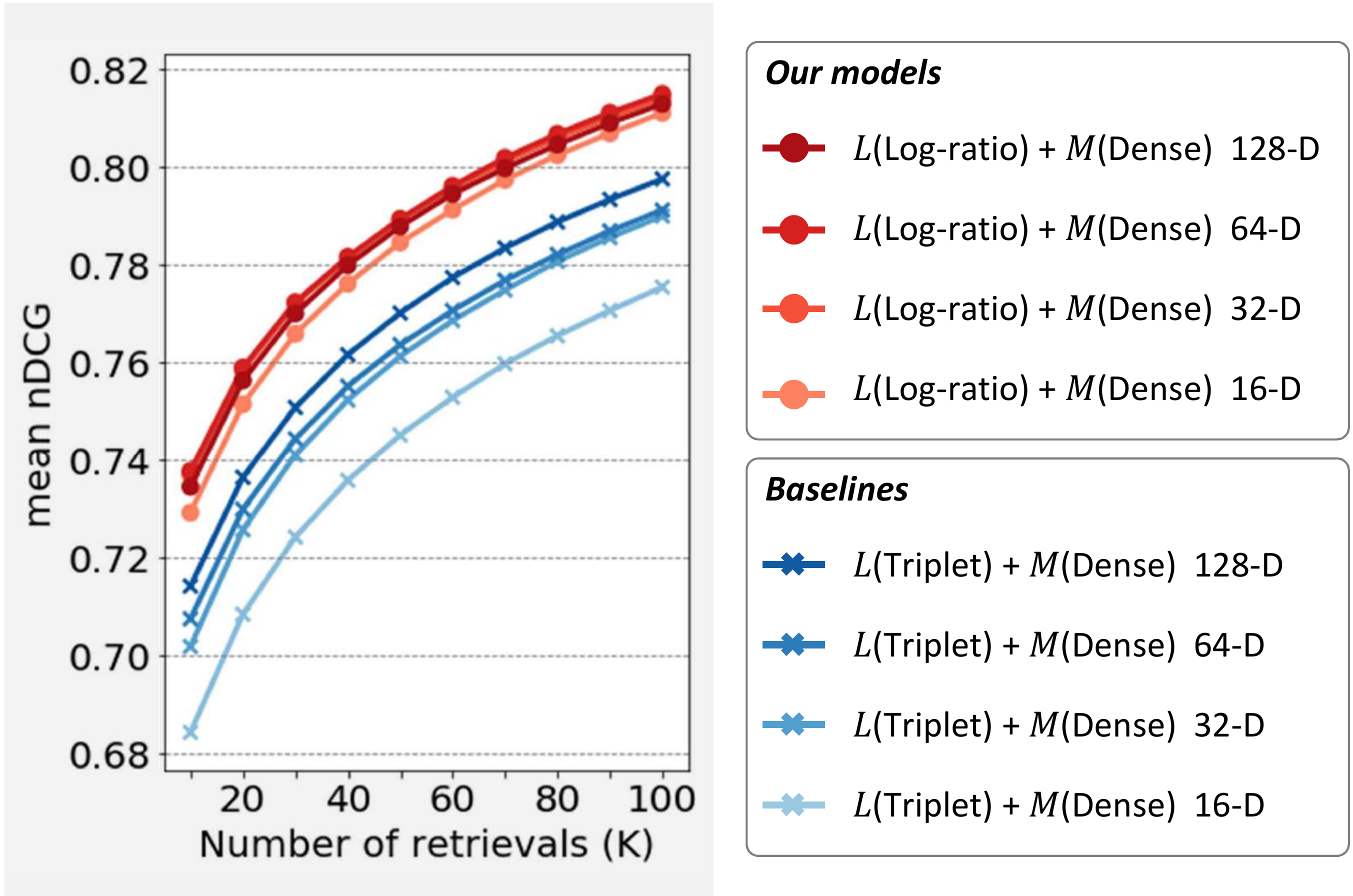}
\caption{
Performance versus embedding dimensionality.
} 
\vspace{-0.2cm}
\label{fig:pose_dimension}
\end{figure}

Finally, we evaluate the human pose retrieval performance by varying embedding dimensionality to show how much effective our embedding space is.
As illustrated in Figure~\ref{fig:pose_dimension}, when decreasing the embedding dimensionality to 16, the performance of our model drops marginally while that of $L$(\textsf{\small Triplet})+$M$(\textsf{\small Dense}) is reduced significantly.
Consequently, the 16-D embedding of our model outperforms 128-D embedding of $L$(\textsf{\small Triplet})+$M$(\textsf{\small Dense}).
This result demonstrates the superior quality of the embedding space learned by our log-ratio loss.

\begin{figure*} [t]
\centering
\includegraphics[width = 0.98 \textwidth]{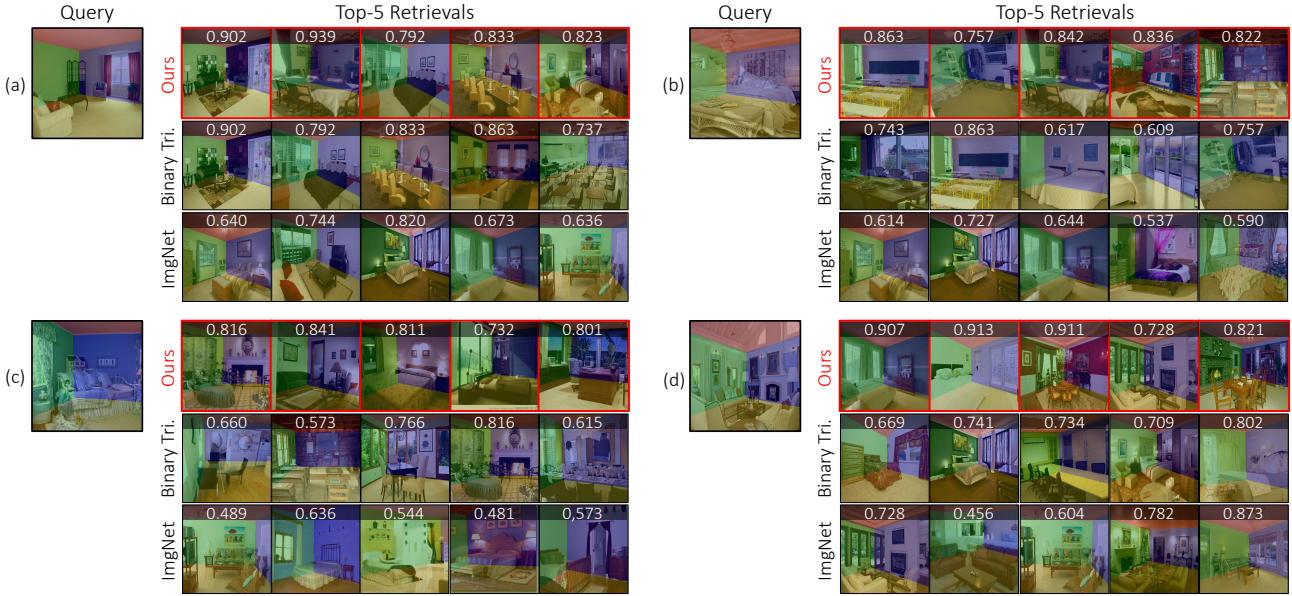}
\caption{
Qualitative results of room layout retrieval. 
For an easier evaluation, the retrieved images are blended with their groundtruth masks, and their mIoU scores are reported together.
\textsf{\footnotesize Binary Tri.}: $L$(\textsf{\footnotesize Triplet})+$M$(\textsf{\footnotesize Binary}). \textsf{\footnotesize ImgNet}: ImageNet pretraiend ResNet101.
} 
\vspace{-0.2cm}
\label{fig:layout_qualitative}
\end{figure*}

\subsection{Room Layout Retrieval}

The goal of this task is to retrieve images whose 3-D room layouts are most similar with that of query image, with no explicit layout estimation in test time.
We define the distance between two rooms $i$ and $j$ in terms of their room layouts as $1-\textrm{mIoU}(R_i, R_j)$, where $R$ denotes the groundtruth room segmentation map and mIoU denotes mean Intersection-over-Union.

Since this paper is the first attempt to tackle the room layout retrieval task, we compare our approach only with the common baselines. 
As shown quantitatively in Figure~\ref{fig:quantitative_eval}(b), the advantage of the dense triplet mining is not significant in this task,
probably because room layout labels of the training images are diverse and sparse so that it is not straightforward to sample triplets densely. 
Nevertheless, our model outperforms all the baselines by a noticeable margin thanks to the effectiveness of our log-ratio loss.

Qualitative results of the room layout retrieval are illustrated in Figure~\ref{fig:layout_qualitative}.
As in the case of the pose retrieval, results of the ImageNet pretrained model are frequently affected by object classes irrelevant to room layouts (\eg, bed in Figure~\ref{fig:layout_qualitative}(b) and sofa in Figure~\ref{fig:layout_qualitative}(d)), while those of our approach are accurate and robust against such distractors.



\begin{figure*} [!t]
\centering
\includegraphics[width = 0.98 \textwidth]{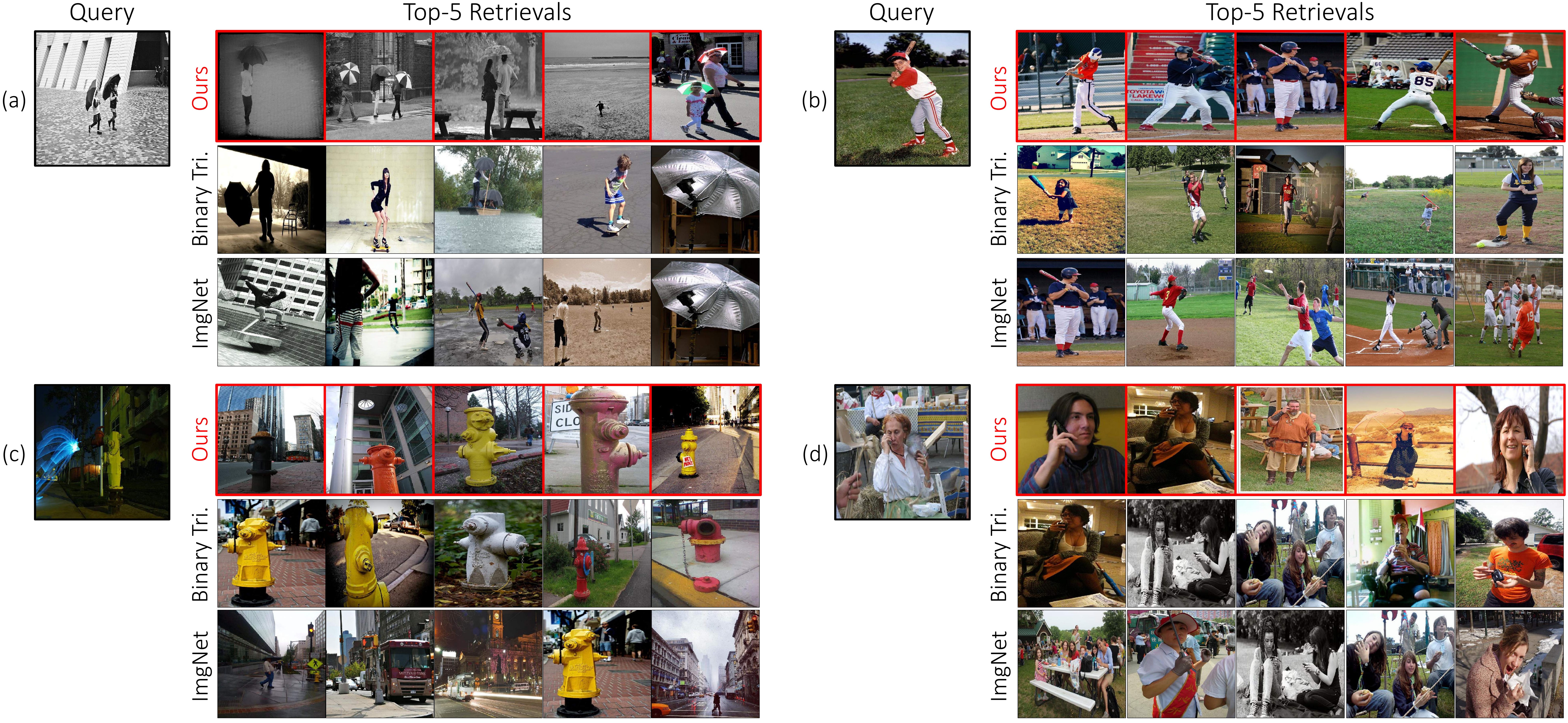}
\caption{
Qualitative results of caption-aware image retrieval. 
\textsf{\footnotesize Binary Tri.}: $L$(\textsf{\footnotesize Triplet})+$M$(\textsf{\footnotesize Binary}). \textsf{\footnotesize ImgNet}: ImageNet pretraiend ResNet101.
} 
\vspace{-0.1cm}
\label{fig:caption_qualitative}
\end{figure*}

\subsection{Caption-aware Image Retrieval}
\label{sec:caption_aware_IR}

    
An image caption describes image content thoroughly.
It is not a simple combination of object classes, but involves richer information including their numbers, actions, interactions, relative locations.
Thus, using caption similarities as supervision allows our model to learn image relations based on comprehensive image understanding.

Motivated by this, we address the caption-aware image retrieval task, which aims to retrieve images described by most similar captions with query.
To define a caption-aware image distance, we adopt a sentence distance metric called Word Mover's Distance (WMD)~\cite{WMD}.
Let $W(x,y)$ be the WMD between two captions $x$ and $y$. 
As each image in our target dataset~\cite{Mscoco} has 5 captions, we compute the distance between two caption sets $X$ and $Y$ through WMD by
\begin{align}
W(X,Y)= \sum_{x\in X} \min_{y\in Y}W(x,y) + \sum_{y\in Y} \min_{x\in X}W(x,y).
\label{eq:wmd}
\end{align}

We train our model and the common baselines with the WMD labels. 
As shown in Figure~\ref{fig:quantitative_eval}(c), our model outperforms all the baselines, and both of the log-ratio loss and the dense triplet mining clearly contribute to the performance boost, while the improvement is moderate due to the difficulty of the task itself.
As illustrated in Figure~\ref{fig:caption_qualitative},
our model successfully retrieves images that contain high-level image content described by queries like object-object interactions (\eg, person-umbrella in Figure~\ref{fig:caption_qualitative}(a)),
object actions (\eg,holding something in Figure~\ref{fig:caption_qualitative}(b,d)),
and specific objects of interest (\eg, hydrant in Figure~\ref{fig:caption_qualitative}(c)).
In contrast, the two baselines in Figure~\ref{fig:caption_qualitative} often fail to retrieve relevant images, especially those for actions and interactions.

\subsection{Representation Learning for Image Captioning}


An ImageNet pretrained CNN has been widely adopted as an initial or fixed visual feature extractor in many image captioning models~\cite{chen2017sca,wang2017diverse,dai2017contrastive,rennie2017self}. 
As shown in Figure~\ref{fig:caption_qualitative}, however, similarities between image pairs in the ImageNet feature space do not guarantee their caption similarities.
One way to further improve image captioning quality would be exploiting caption labels for learning a visual representation specialized to image captioning.

We are motivated by the above observation, and believe that a CNN learned with caption similarities through our continuous metric learning framework can be a way to implement the idea.
To this end, we adopt our caption-aware retrieval model described in Section~\ref{sec:caption_aware_IR} as an initial, caption-aware visual feature extractor of two image captioning networks: Att2all2~\cite{rennie2017self} and Topdown~\cite{anderson2018bottom}.
Specifically, our caption-aware feature extractor is compared with the ImageNet pretrained baseline of ours, and $(14\times14\times2048)$ average pooled outputs of their last convolution layers are utilized as caption-aware and ImageNet pretrained features.
For training the two captioning networks, we directly follow the training scheme proposed in~\cite{rennie2017self}, which first pretrains the networks with cross-entropy (XE) loss then finetunes them using reinforcement learning (RL) with the CIDEr-D~\cite{CIDEr} metric.

\begin{table}[t]
\resizebox{0.49\textwidth}{!}{
\begin{tabular}{c|ll|ccccc}
\hline
Model                     & \multicolumn{2}{c|}{Train}     & B4 & C & M & R & S \\ \hline
\multirow{4}{*}{ATT} & \multirow{2}{*}{Img}                & XE    & 0.3302	& 1.029	 & 0.2585 &	0.5456 & 0.192  \\ 
                          &                                & RL    & 0.3348 & 1.131  & 0.2630  & 0.5565 & 0.1965 \\ \cline{2-8}
                          & \multirow{2}{*}{Cap}           & XE    & 0.3402	& 1.052  & 0.2608 & 0.5504 & 0.1942 \\ 
                          &                                & RL    & \textbf{0.3465}	& \textbf{1.159}	 & \textbf{0.2673} & \textbf{0.5613} & \textbf{0.2010}  \\ \cline{2-8} 
                          \hline
\multirow{4}{*}{TD}  & \multirow{2}{*}{Img}                & XE    & 0.3421 & 1.087 & 0.2691 & 0.5543 & 0.2011  \\ 
                          &                                & RL    & 0.3573	& 1.201	& 0.2739 & 0.5699 & 0.2085  \\ \cline{2-8}
                          & \multirow{2}{*}{Cap}           & XE    & 0.3479	& 1.097	& 0.2707 & 0.5573 & 0.2012  \\ 
                          &                                & RL    & \textbf{0.3623}	& \textbf{1.213}	& \textbf{0.2758} & \textbf{0.5718} & \textbf{0.2107}  \\ \cline{2-8} 
                        \hline
\end{tabular}
}
\\
\caption{Captioning performance on the Karpathy test split~\cite{Karpathy_mscoco_split}. 
We report scores obtained by a single model with the beam search algorithm (beam size $=2$).
ATT: Att2all2~\cite{rennie2017self}.
TD: Topdown~\cite{anderson2018bottom}. 
Img: ImageNet pretrained feature.
Cap: Caption-aware feature.
XE: Pretrained with cross-entropy.
RL: Finetuned by reinforcement learning.
B4: BLEU-4~\cite{papineni2002bleu}.
C: CIDEr-D~\cite{CIDEr}.
M: METEOR~\cite{denkowski2014meteor}.
R: ROGUE-L~\cite{lin2004rouge}.
S: SPICE~\cite{anderson2016spice}.}
\vspace{-0.1cm}
\label{tab:caption_quantitative}
\end{table}

\begin{figure} [t]
\centering
\includegraphics[width = 1\columnwidth]{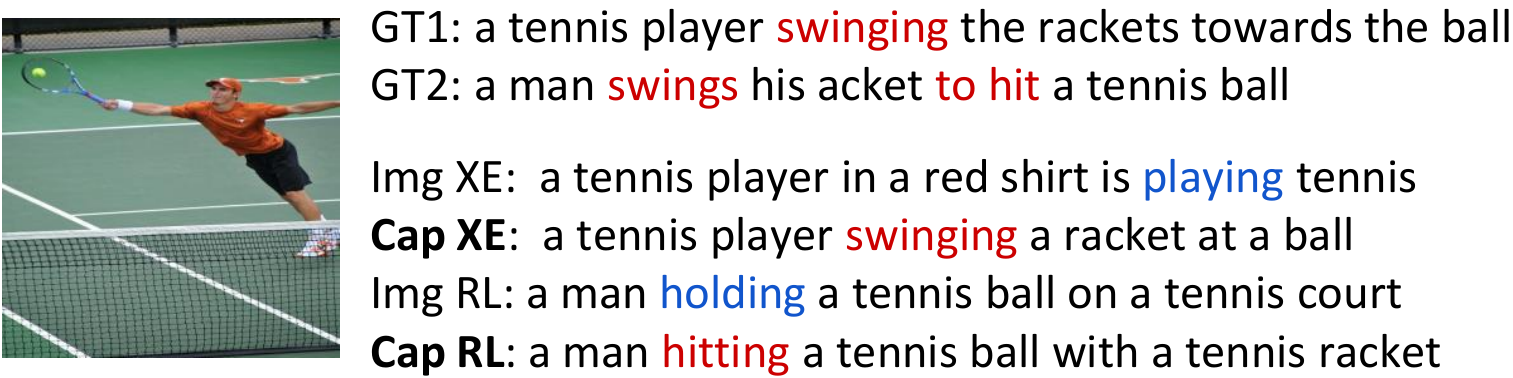}
\caption{Captions generated by the Topdown attention~\cite{anderson2018bottom}.
GT: groundtruth caption.
Img: ImageNet pretrained feature.
Cap: Caption-aware feature.
XE: Pretrained with cross-entropy.
RL: Finetuned by reinforcement learning.}
\vspace{-0.1cm}
\label{fig:caption_generated}
\end{figure}

Table~\ref{tab:caption_quantitative} quantitatively summarizes captioning performance of the ImageNet pretrained feature and our caption-aware feature. 
The scores of reproduced baseline are similar or higher than those reported in its original paper. 
Nonetheless, our caption-aware feature consistently outperforms the baseline in all evaluation metrics and for both of two captioning models.
Also, qualitative examples of captions generated by the models in Table~\ref{tab:caption_quantitative} are presented in Figure~\ref{fig:caption_generated}, where baselines generate incorrect captions while the models based on our caption-aware feature avoid choosing the wrong word and generate better captions.

\section{Conclusion}
\label{sec:conclusion}
We have presented a novel loss and tuple mining strategy for deep metric learning using continuous labels. 
Our approach has achieved impressive performance on three different image retrieval tasks with continuous labels using human poses, room layouts and image captions. 
Moreover, we have shown that our framework can be used to learn visual representation with continuous labels. 
In the future, we will explore the effect of label distance metrics and a hard tuple mining technique for continuous metric learning to further improve the quality of learned metric space.

\vspace{0.3cm}
{\small
\noindent \textbf{Acknowledgement:} This work was supported by Basic Science Research Program and R\&D program for Advanced Integrated-intelligence for IDentification  through the National Research Foundation of Korea funded by the Ministry of Science, ICT (NRF-2018R1C1B6001223, NRF-2018R1A5A1060031, NRF-2018M3E3A1057306,
NRF-2017R1E1A1A01077999), 
and by the Louis
Vuitton - ENS Chair on Artificial Intelligence.
}


\newpage
{\small
\bibliographystyle{ieee}
\bibliography{metric_SYMK_cvpr19}
}

\renewcommand\thesection{\Alph{section}}
\setcounter{section}{0}

\section{Appendix}
In this appendix, we first present more qualitative results omitted from the main paper due to space limit. 
Results for human pose retrieval, room layout retrieval, and caption-aware image retrieval are given in Section~\ref{sec:pose_retrieval}, Section~\ref{sec:layout_retrieval}, and Section~\ref{sec:caption_retrieval}, respectively.
Also, Section~\ref{sec:captioning} provides implementation details, more qualitative examples, and in-depth analysis for image captioning with our caption-aware visual features.





\subsection{Human Pose Retrieval}
\label{sec:pose_retrieval}
\begin{figure*} [t]
\centering
\includegraphics[width = 1 \textwidth]{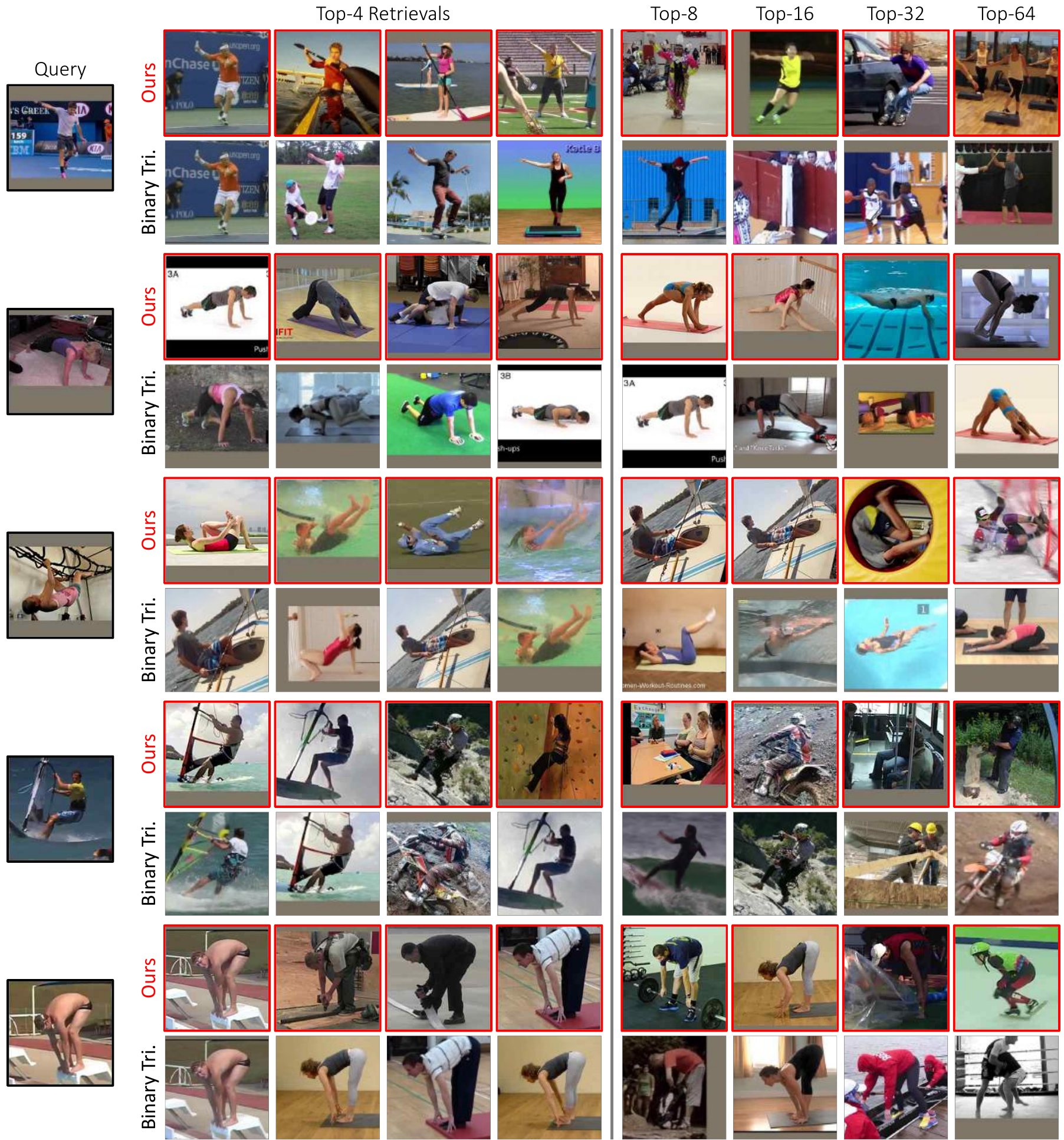}
\caption{
Qualitative results of human pose retrieval on the MPII human pose dataset.
\textsf{\footnotesize Ours}: $L$(\textsf{\footnotesize Log-ratio})+$M$(\textsf{\footnotesize Dense}).
\textsf{\footnotesize Binary Tri.}: $L$(\textsf{\footnotesize Triplet})+$M$(\textsf{\footnotesize Binary}).
} 
\label{fig:pose_sup_qualitative}
\end{figure*}

More qualitative examples for human pose retrieval are presented in Figure~\ref{fig:pose_sup_qualitative}.
As in the main paper, results of our model are compared with those of a  baseline $L$(\textsf{\small Triplet})+$M$(\textsf{\small Binary}), which is an advanced version of the previous approach~\cite{thin_slicing} adopting the same backbone network with ours for a fair comparison.
From overall results, one can observe that even top 64 retrievals are relevant to queries in the results of our model, while those of the baseline are sometimes incorrect 
when a large number of images are retrieved.
Also, when the query exhibits a rare pose as in the 2nd and 4th rows, the baseline includes irrelevant images (\eg, horizontally flipped poses) even among top-4 retrievals, while ours still works successfully.

\begin{figure*} [t]
\centering
\includegraphics[width = 1 \textwidth]{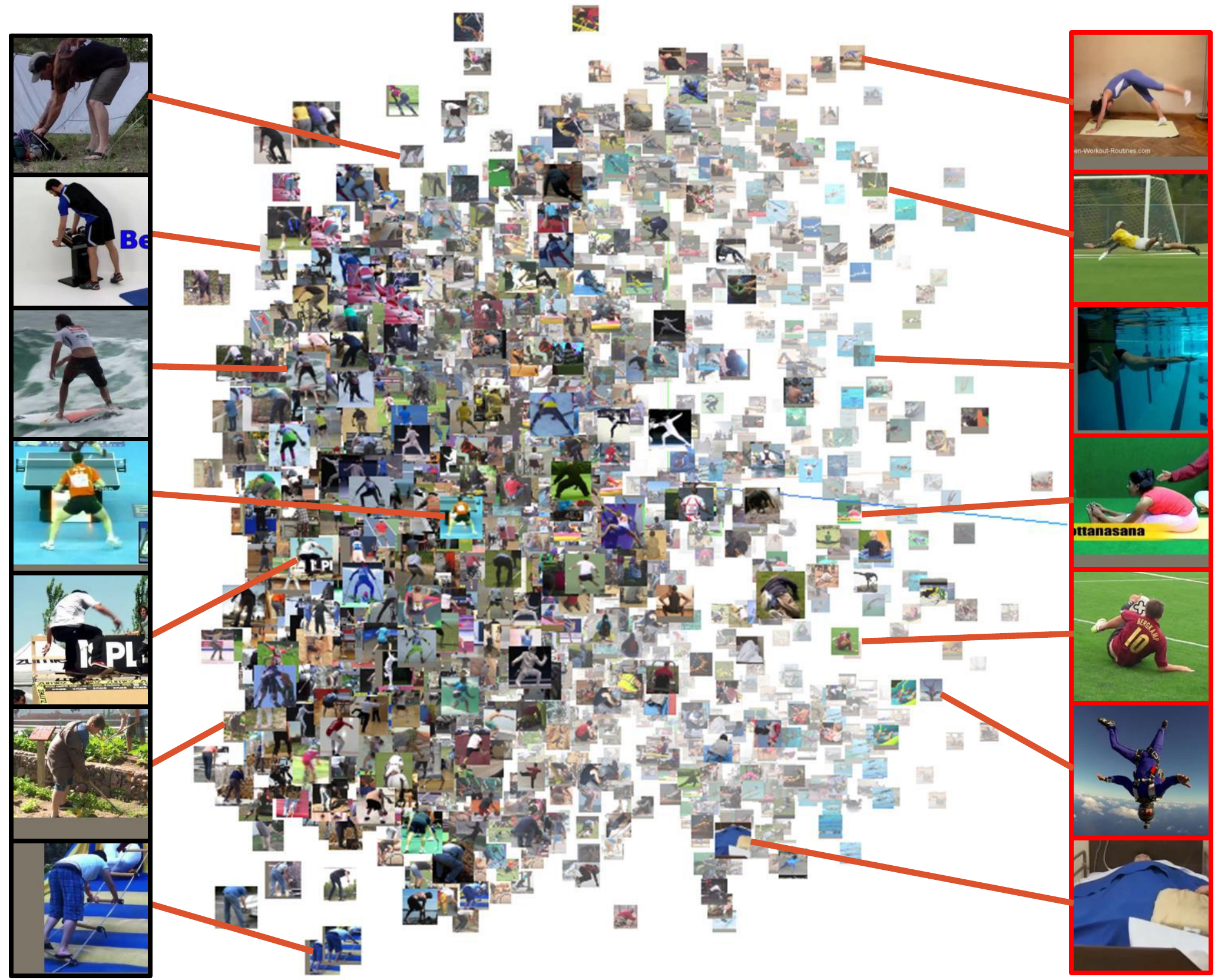}
\caption{
3D t-SNE visualization of our pose embedding manifold learned on the MPII human pose dataset. 
Images with black boundaries indicate common poses and those with red boundaries exhibit rare poses.
} 
\label{fig:pose_tsne}
\end{figure*}

In Figure~\ref{fig:pose_tsne}, we visualize the embedding manifold learned by our approach through 3D t-SNE.
In the manifold, poses are locally consistent and change smoothly between two distant coordinates as illustrated by the series of pose images on the left-hand side of Figure~\ref{fig:pose_tsne}.
More interestingly, common poses (images with black boundaries) are densely populated while rare ones (images red boundaries) are scattered on the other side.
This observation implicitly indicates that the learned manifold preserves degrees of similarities between images.


\subsection{Room Layout Retrieval}
\label{sec:layout_retrieval}
\begin{figure*} [t]
\centering
\includegraphics[width = 1 \textwidth]{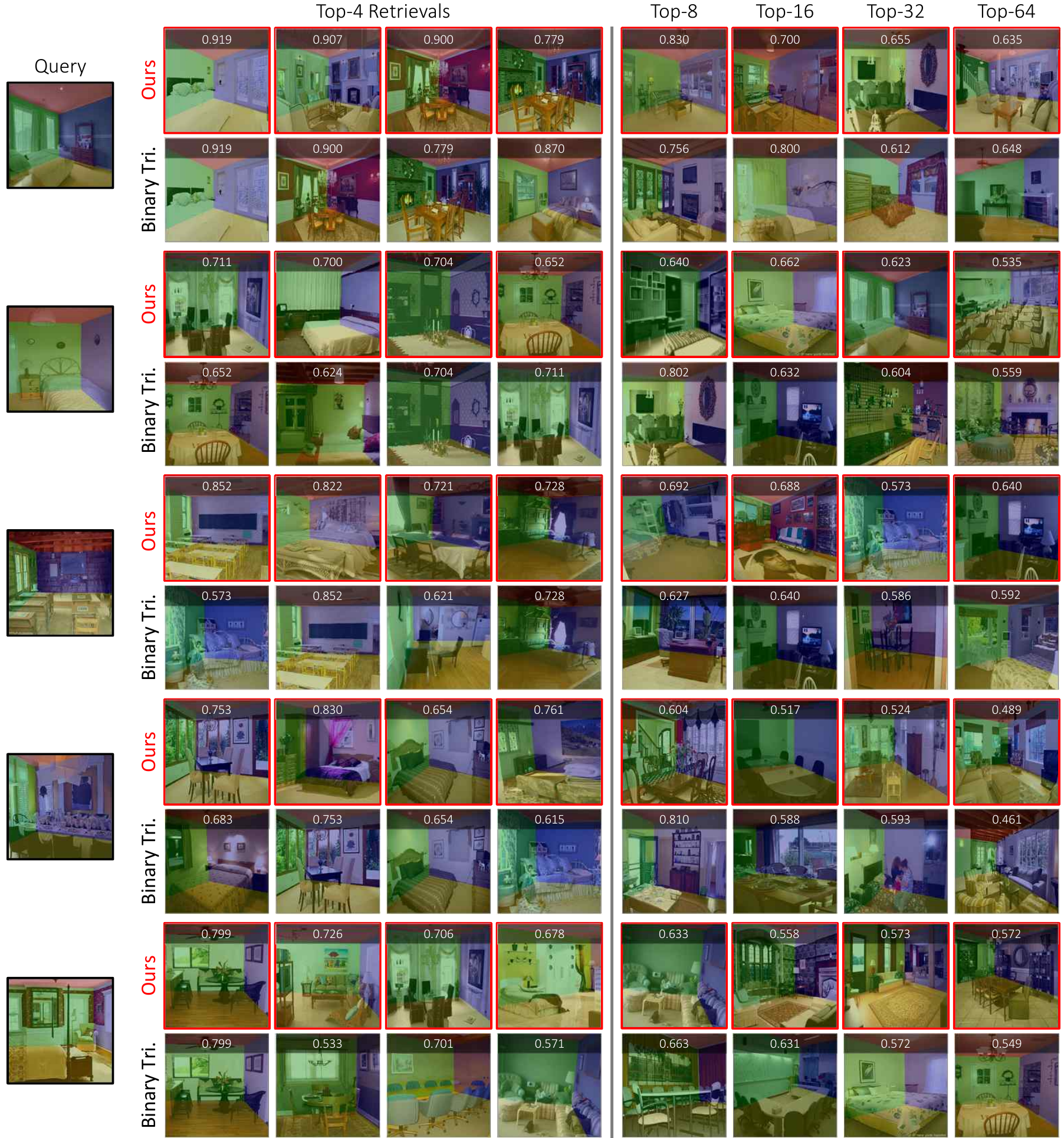}
\caption{
Qualitative results of room layout retrieval on the LSUN room layout dataset.
\textsf{\footnotesize Ours}: $L$(\textsf{\footnotesize Log-ratio})+$M$(\textsf{\footnotesize Dense}).
\textsf{\footnotesize Binary Tri.}: $L$(\textsf{\footnotesize Triplet})+$M$(\textsf{\footnotesize Binary}).
} 
\label{fig:layout_sup_qualitative}
\end{figure*}

More qualitative examples of retrieved room layouts are illustrated in Figure~\ref{fig:layout_sup_qualitative}. The images are blended with their groundtruth segmentation masks. The values above the images are the mIOU scores with the query. Noted that the higher the mIOU score, the more relevant to the query is.
As in Section~\ref{sec:pose_retrieval}, our model is compared with a baseline $L$(\textsf{\small Triplet})+$M$(\textsf{\small Binary}).

From overall results, all images retrieved as top1 by our model have higher mIOU values than those of the baseline. Also, one can observe that our top 4 retrieval results have a higher mIOU than the results of baseline. As in the 1st, 2nd and 4th rows, baseline has images that among top-4 retrievals are less relevant than top 8 or top 16 retrievals to the queries 
On the other hand, the results show that as our method retrieves a large number of images, images gradually have a smaller value of mIOU as common sense. Overall, our model shows better performance than the baseline.

\subsection{Caption-aware Image Retrieval}
\label{sec:caption_retrieval}
\begin{figure*} [t]
\centering
\includegraphics[width = 1 \textwidth]{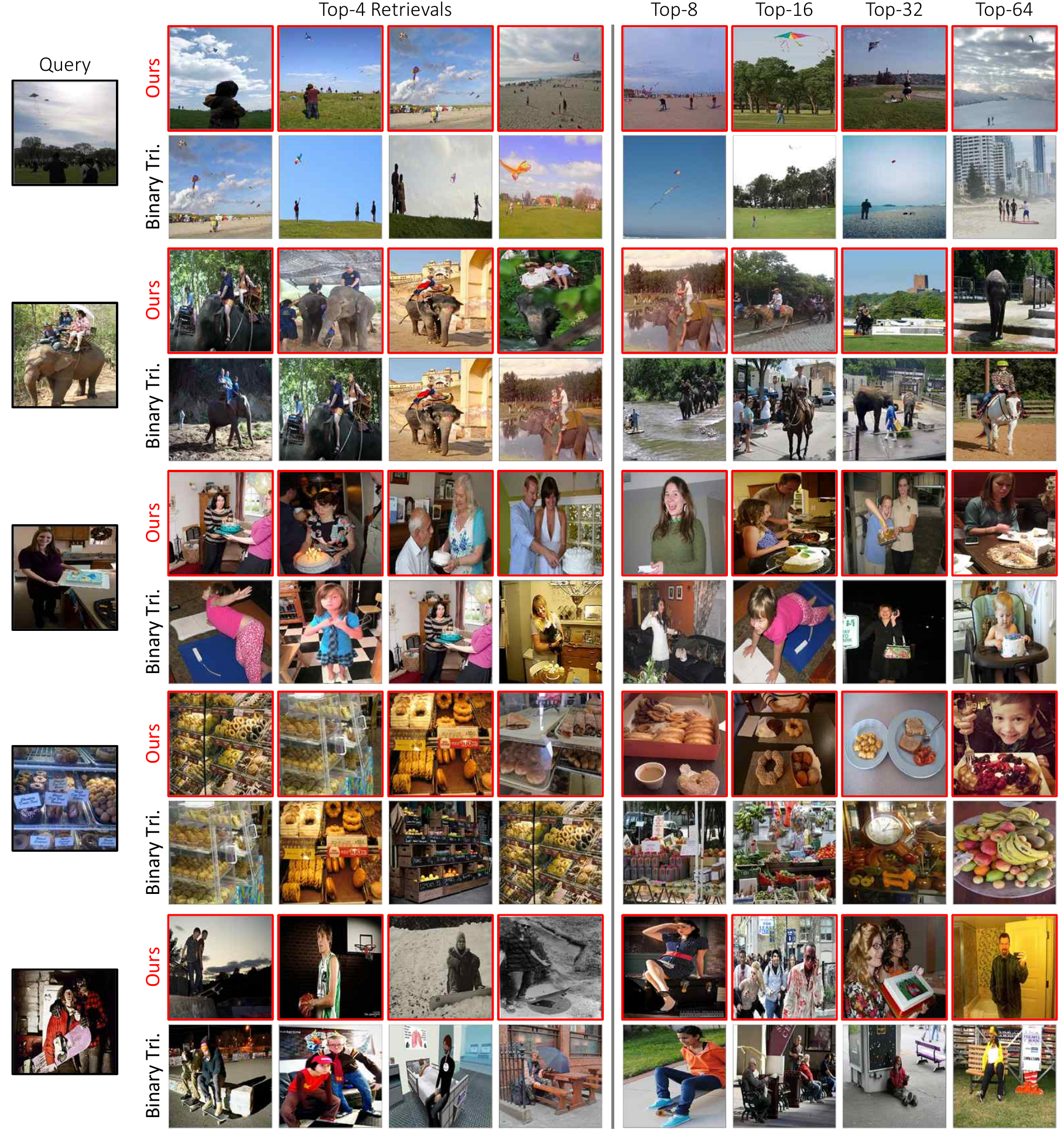}
\caption{
Qualitative results of caption-aware image retrieval on the MS-COCO dataset.
\textsf{\footnotesize Ours}: $L$(\textsf{\footnotesize Log-ratio})+$M$(\textsf{\footnotesize Dense}).
\textsf{\footnotesize Binary Tri.}: $L$(\textsf{\footnotesize Triplet})+$M$(\textsf{\footnotesize Binary}).
} 
\label{fig:caption_sup_qualitative}
\end{figure*}

Our model and a baseline $L$(\textsf{\small Triplet})+$M$(\textsf{\small Binary}) are compared qualitatively with more retrieval examples in Figure~\ref{fig:caption_sup_qualitative}.
Note that the baseline can be considered as a variant of the existing caption-aware retrieval model~\cite{Gordo_cvpr2017}, which also employs a triplet loss and a tuple mining strategy based on nearest neighbors as the baseline does.

From 1st to 3rd rows are images of people who take specific actions. (e.g., someone holding a cake) And the 4th rows are images of the object at a particular location. (e.g., donuts in a display case)
One may notice that the retrieval results can be less relevant after top 16 since some images have a few relevant examples.
For the 1st and 2nd queries, both models successfully retrieves relevant images. On the other hand, in the 3rd and 4th rows of the figure, our model performs better than the baseline. In the 3rd row, the baseline sometimes retrieves images that are not related to the 'cake' at all. In the 4th row, our model retrieves images related to the 'donut' even to the top 16, but one can see that the baseline does not work well. For the last query, both models fail to retrieve relevant images since the query image is unusual and the number of testing images relevant to the query is not sufficiently large.


\subsection{Image Captioning}
\label{sec:captioning}

Our experiments are based on the implementation of \cite{luo2018discriminability}.
For att2all2 model, our hyperparameters are slightly different in training epoch, batch size, and learning rate during reinforcement learning from those of~\cite{rennie2017self}. 
We first train the captioning model using cross entropy loss with batch size of 16 until 20 epoch and then increase the batch size to 32, set learning rate 5e-4 and further 30 epoch of reinforcement learning.
For topdown model, hyperparameters are same as in att2all2 model except batch size of 32 and the number of hidden units of 1024 for each LSTM.
Note that this setting is different from the original work of topdown model \cite{anderson2018bottom}



\begin{figure*}
    \centering
    \includegraphics[width = 1\textwidth]{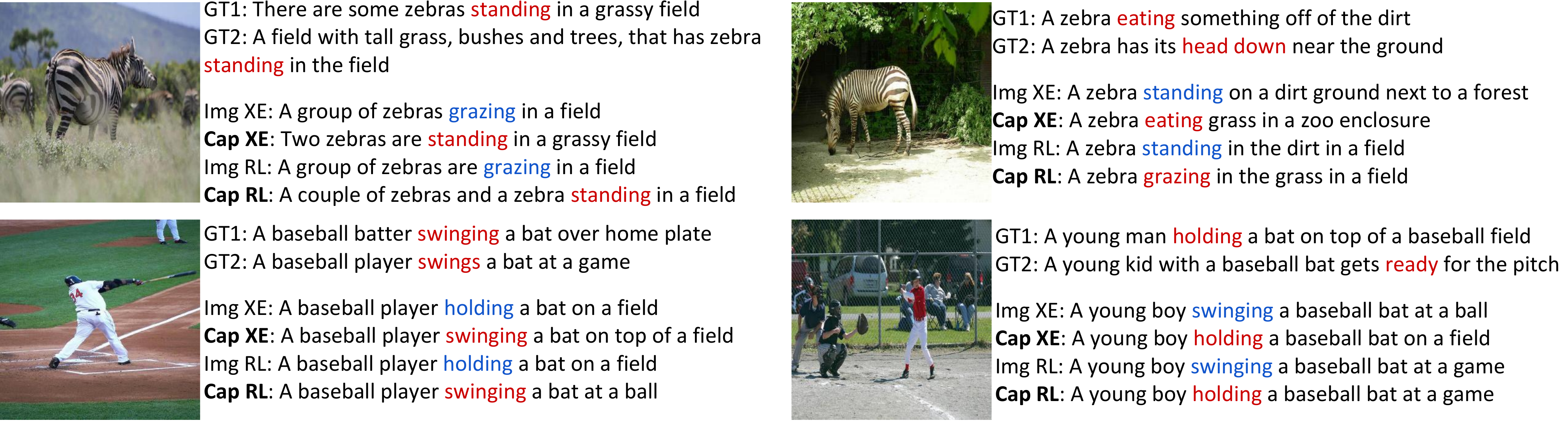}
    \caption{Two pairs of complementary examples from topdown model.
    GT: Groundtruth caption. Img: ImageNet pretrained feature. Cap: Caption-aware feature. XE: Pretrained with cross-entropy. RL: Finetuned by reinforcement learning}
    \label{fig:cap_gen1}
    \medskip
    \centering
    \includegraphics[width = 1\textwidth]{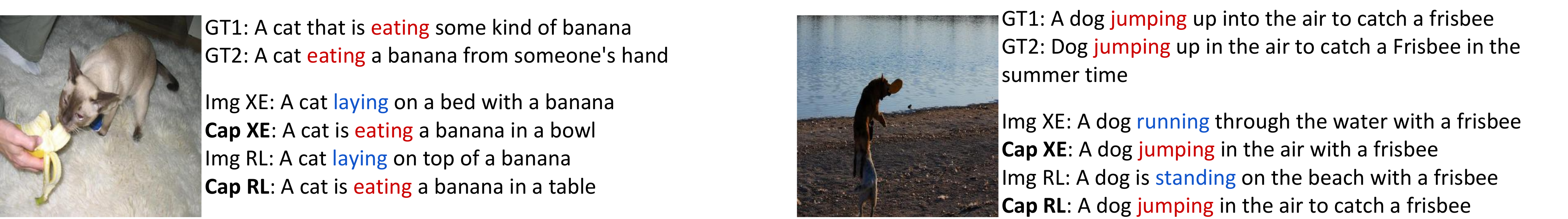}
    \caption{Two examples from att2all2 model}
    \label{fig:cap_gen2}
    \medskip
    \centering
    \includegraphics[width = 1\textwidth]{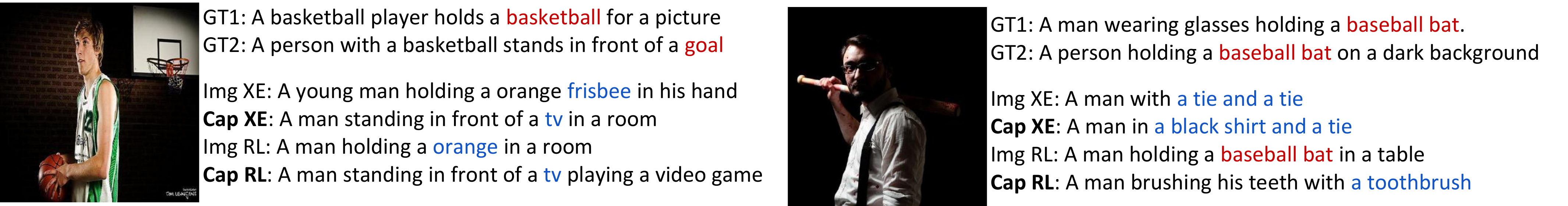}
    \caption{Two failure cases from topdown model}
    \label{fig:cap_gen3}
\end{figure*}





We are going to introduce six interesting examples in figure \ref{fig:cap_gen1}, figure \ref{fig:cap_gen2} and two failure cases in figure \ref{fig:cap_gen3}. 
Figure \ref{fig:cap_gen1} includes two complementary pairs that the model learning from ImageNet pretrained features confuses action for each other.
On the other hand, the model learning with from caption aware feature generates the sentences without confusion.
This result seems to be that the model distinguish the visual semantic properly rather than repeating one side of the complementary pairs.
Figure \ref{fig:cap_gen2} shows two examples from att2all2 model. 
These examples show that both models (att2all2, topdown model) have similar characteristics.
Two failure cases of our feature are \ref{fig:cap_gen3}.
In both cases, model with our feature fails to produce a proper caption due to find the object correctly.


\begin{figure*}
    \centering
    \includegraphics[width = \textwidth]{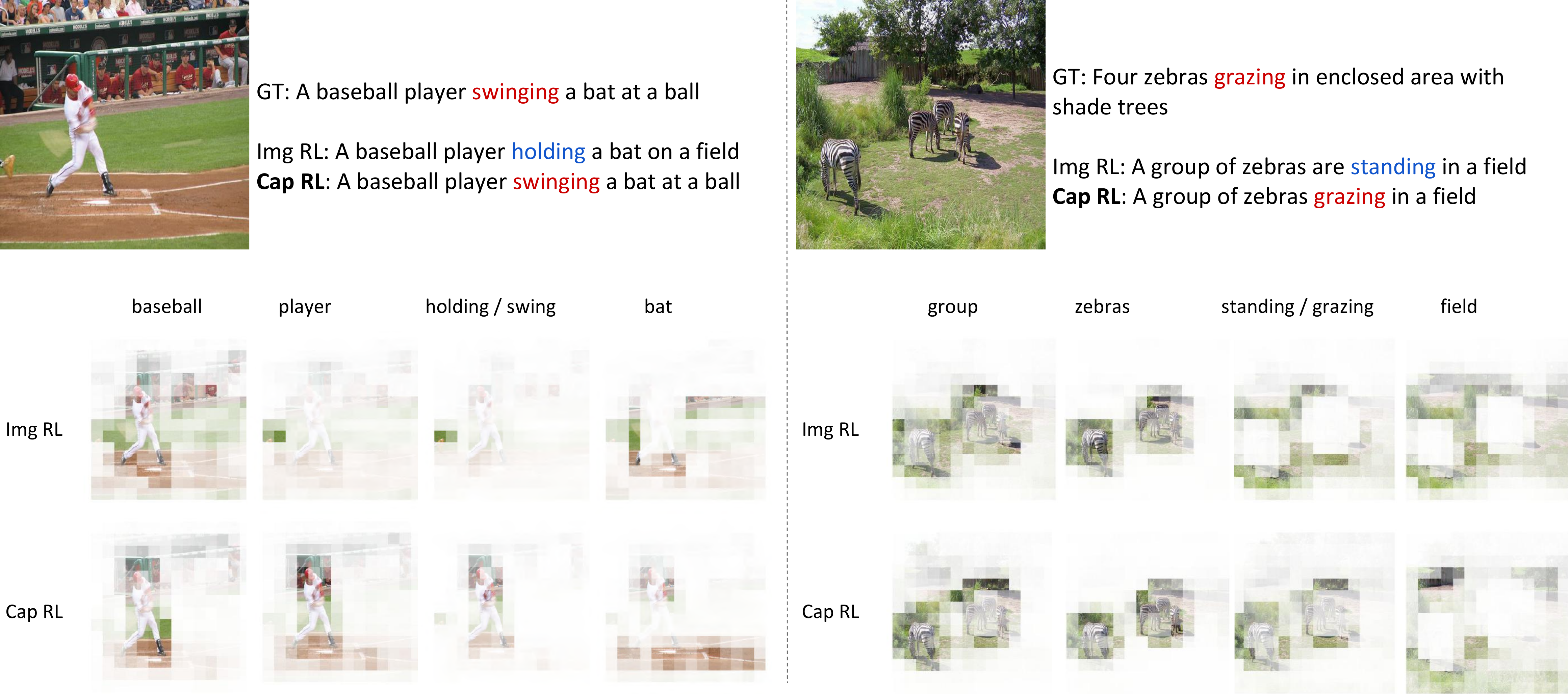}
    \caption{Attention maps of typical examples of from reinforcement learned att2all2 model with ImageNet pretrained feature (Img RL), caption aware feature(Cap RL)}
    \label{fig:cap_att}
\end{figure*}

%

At figure \ref{fig:cap_att} we try to check the difference of attention in the two typical examples.
For these examples, each model using Imagenet pretrained feature, caption aware feature generates almost same caption except behavior.
However, the tendency of attention is slightly different between two models.
In the case of baseline model, strong attention is given to marginal part of image while predict the action.
Whereas, the object took strong attention from the model using our feature.
This result correspond to our common sense that we need to focus on object in order to notice subtle changes in posture.


\end{document}